\documentclass[lettersize,journal]{IEEEtran}
\usepackage{amsmath,amsfonts}
\usepackage{array}
\usepackage[caption=false,font=normalsize,labelfont=sf,textfont=sf]{subfig}
\usepackage{textcomp}
\usepackage{stfloats}
\usepackage{url}
\usepackage{verbatim}
\usepackage{graphicx}
\usepackage{cite}
\usepackage{makecell}

\usepackage[hidelinks]{hyperref}
\usepackage[ruled,vlined]{algorithm2e}
\usepackage{xcolor}
\usepackage{enumitem}

\usepackage{booktabs}
\usepackage{multirow}
\usepackage{colortbl}

\definecolor{AEMRow}{HTML}{FAF7FF}
\definecolor{ImproveRed}{HTML}{C00000}

\newcommand{\best}[1]{\textbf{#1}}
\newcommand{\gain}[1]{\textcolor{ImproveRed}{\scriptsize{(#1)}}}

\definecolor{AEMPurple}{HTML}{3B168A}
\definecolor{AEMBlue}{HTML}{3B73C5}
\definecolor{AEMOrange}{HTML}{D2691E}
\definecolor{AEMGreen}{HTML}{2E7D32}

\newcommand{\vis}[1]{\textcolor{AEMBlue}{#1}}
\newcommand{\act}[1]{\textcolor{AEMOrange}{#1}}
\newcommand{\mem}[1]{\textcolor{AEMGreen}{#1}}
\newcommand{\stage}[2]{\noindent\textcolor{AEMPurple}{\textbf{#1. #2}}}

\SetKwInput{KwInput}{Input}
\SetKwInput{KwOutput}{Output}
\SetKwComment{Comment}{$\triangleright$\ }{}

\begin{document}

\title{Action-Effect Memory Pretraining for Robot Manipulation}

\author{
\textbf{Yijing Zhou}\textsuperscript{1,2,$\ast$}, 
\textbf{Qiwei Liang}\textsuperscript{1,2,$\ast$,$\ddagger$}, 
\textbf{Sitong Zhuang}\textsuperscript{2}, 
\textbf{Jiaxi Li}\textsuperscript{2}, \\
\textbf{Xianpeng Wang}\textsuperscript{1},
\textbf{Boyang Cai}\textsuperscript{1,2},
\textbf{Yunyang Mo}\textsuperscript{1},  
\textbf{Renjing Xu}\textsuperscript{1,$\dagger$} \\
\normalsize

\textsuperscript{1}Hong Kong University of Science and Technology (Guangzhou) 
\textsuperscript{2}Shenzhen University \quad \\
\normalsize
$^\ast$Equal contribution \quad 
$^\dagger$Corresponding author \quad
$^\ddagger$Project leader \\
}




\maketitle

\begin{abstract}
We present \textbf{AEM}, an \textbf{Action-Effect Memory} pretraining framework for robot manipulation that learns compact temporal representations from vision-action history. Unlike prior robot representation pretraining methods that mainly focus on single-frame visual encoding, AEM targets the temporal nature of manipulation, where the current observation alone is often insufficient under partial observability. AEM models manipulation as an action-driven interaction process by interleaving visual and action features and applying masked modeling to recover missing content from incomplete histories, thereby learning action-conditioned state evolution. The Mamba-encoded output of the final vision token is used as a compact history representation, serving as the global context for decoding and downstream control. This design preserves a single-vector temporal bottleneck while keeping inference efficient. We evaluate AEM with \textbf{Diffusion Policy} and \textbf{Flow Policy}. AEM consistently improves manipulation performance in both simulation and real-world settings, outperforming baselines across clean scenes, cluttered and random scenes, and non-Markovian tasks. Ablation studies further show that history-aware pretraining surpasses single-frame pretraining and direct frame stacking, while reducing inference latency and computational cost. Project homepage: \url{https://dongqiuyijing.github.io/AEM-Research-Homepage-latest/}
\end{abstract}

\begin{IEEEkeywords}
Robot Manipulation, Temporal Representation Pretraining, Action-Effect Memory.
\end{IEEEkeywords}

\section{Introduction}
\IEEEPARstart{R}{epresentation} pretraining for robot manipulation has advanced rapidly in recent years~\cite{dynamo,srirama2024hrp}. Most existing methods primarily focus on single-frame visual encoding, improving spatial representations of the current observation through masked reconstruction~\cite{3d-mvp,ma2022vip,sensorimotor}, contrastive learning~\cite{r3m_ref,RPR}, and geometric consistency constraints~\cite{spa}. These approaches have significantly improved static scene understanding and promoted the broad adoption of visual pretraining in robot manipulation. However, they still formulate pretraining largely as a single-timestep observation encoding problem, paying limited attention to the temporal information that is often crucial for manipulation~\cite{zhou2025mtil}.

This limitation stems from the sequential nature of robot manipulation. In real-world settings, robots often face occlusion, contact changes, short-term partial observability, and delayed environmental responses, making the current observation insufficient to characterize the state~\cite{torne2026mem,intelligence2026pi}. Historical information is therefore essential for mitigating partial observability, inferring state evolution, and enabling smooth control. A straightforward way to exploit history is to stack past observations or features and feed them into the policy~\cite{lin2026hif}. While effective in some cases, this strategy increases computation, memory usage, and inference latency with history length, and leaves temporal abstraction entirely to the downstream policy. More importantly, raw history stacking does not answer a more fundamental question: \textbf{what kind of memory representation should robot manipulation actually use?}

We argue that an effective memory representation for robot manipulation should satisfy three properties. First, it should be built from vision-action interaction history rather than visual observations alone, since manipulation is driven by actions that change the world~\cite{afro}. Second, it should be dense and compact: its size should not grow with the history horizon, but instead compress interaction histories into a fixed-size representation that can be injected into downstream policies with minimal interface cost. Third, it should capture the structure of interaction across time, especially the dependency between actions and state transitions, so that the resulting memory is informative and usable for robot manipulation.

\begin{figure}[t]
    \centering
    \includegraphics[width=0.95\linewidth]{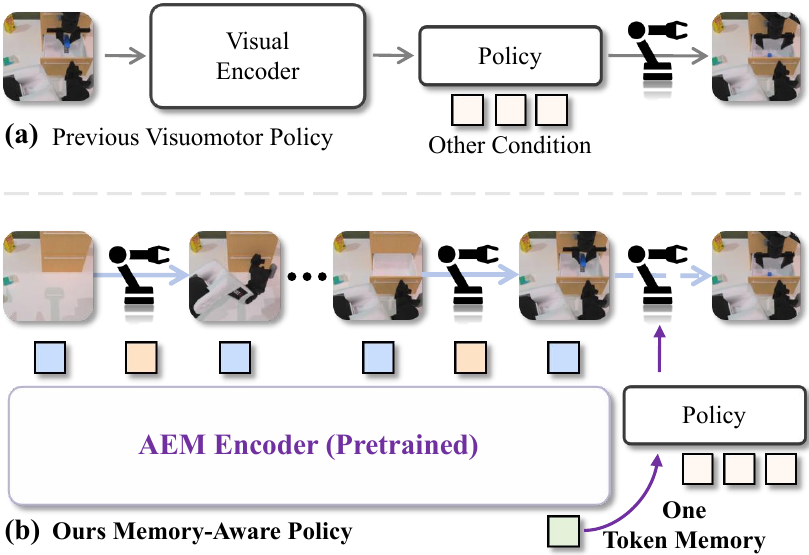}
    \caption{(a) Conventional visuomotor policies encode only the current visual observation and feed the resulting feature into the policy. (b) Our history-aware policy instead pretrains a compact history representation from vision-action interaction history by reusing the encoded final vision token, which is injected into the policy as temporal context. This design provides dense temporal information without introducing a dedicated memory token.}
    \label{fig1:teaser}
\end{figure}

Motivated by these observations, we propose \textbf{AEM}, an \textbf{A}ction-\textbf{E}ffect \textbf{M}emory pretraining framework for robot manipulation. \textbf{AEM} learns a compact history state from long-horizon vision-action interaction history during pretraining and injects it into downstream policies as temporal context. Concretely, we interleave visual and action features in temporal order and pretrain the model to encode the interaction history into a compact latent state representation.

A key challenge is how to make the history representation compact. Many common sequence architectures, especially token-preserving models such as Transformers~\cite{vaswani2017attention}, naturally produce outputs whose token length scales with the input, making them less suitable for distilling histories into a fixed-size dense representation. Instead of appending a learnable memory token, \textbf{AEM} reuses the final vision token in the interleaved history as the compression anchor and takes its Mamba-encoded output as the compact history representation. In this design, Mamba~\cite{gu2023mamba} provides contextual sequence modeling, while the final vision token serves as a fixed-size information bottleneck for history compression.

Another challenge is how to ensure that the compact history representation is not merely stored, but structured and usable. Pure compression objectives may preserve information without guaranteeing that the representation captures the temporal dependencies needed for control. To address this, \textbf{AEM} adopts a masked autoencoding~\cite{he2022masked} objective over interleaved vision-action sequences. By requiring the model to recover missing content from incomplete histories, the final-token representation must encode dependencies across time, including how actions relate to subsequent states. Mamba thus provides the contextual history carrier, while masked autoencoding shapes that carrier into an action-effect representation.

After pretraining, the encoder compresses the entire vision-action history into the final vision token, which can be directly plugged into downstream policies. Temporal information is thus provided to the policy not as raw stacked observations, but as a pretrained compact history representation.

We evaluate \textbf{AEM} on the RoboTwin2.0~\cite{chen2025robotwin} benchmark by integrating it into Diffusion Policy~\cite{chi2025diffusion} and ManiFlow Policy~\cite{yan2025maniflow}. Without introducing a dedicated memory token, \textbf{AEM} consistently improves manipulation performance in both clean and random settings, demonstrating the effectiveness of its final-token history representation. Ablation studies further show that history-aware pretraining yields larger gains than single-frame pretraining and direct frame stacking, while also achieving substantially lower inference latency and computational cost. We also evaluate \textbf{AEM} on RMBench~\cite{chen2026rmbench}, where AEM-enhanced policies achieve higher success rates than the corresponding baselines on challenging non-Markovian tasks, validating its ability to capture useful temporal dependencies beyond the current observation. We further validate \textbf{AEM} on real robots with Diffusion Policy, and the results confirm its effectiveness in real-world manipulation. 
The main contributions of this paper are as follows:
\begin{enumerate}
    \item We propose a temporal representation pretraining paradigm for robot manipulation that compresses long-horizon vision-action interaction history into a single compact final-token representation, providing transferable temporal representations for downstream policies.
    \item We introduce \textbf{AEM}, which combines Mamba-based compact state modeling with masked autoencoding over interleaved vision-action sequences to learn dense and interaction-aware action-effect history representations.
    \item Experiments on RoboTwin2.0 and real robots show that \textbf{AEM} significantly improves manipulation performance without introducing a dedicated memory token. Compared with single-frame pretraining and frame stacking, \textbf{AEM} achieves better performance while reducing inference latency, computational cost, and policy training time.
\end{enumerate}

\section{Related Work}

\textbf{Representation Pretraining for Robot Manipulation.}
Representation pretraining has become an effective way to improve the sample efficiency and generalization of robot manipulation~\cite{shang2024theia,lepert2025masquerade}. Early work mainly transferred general visual pretraining from computer vision to robotics, following either masked reconstruction objectives~\cite{mvp_ref,3d-mvp,radosavovic2023real} similar to MAE or contrastive learning paradigms~\cite{r3m_ref,RPR}. Later work explored richer supervision, such as multi-task learning and robotics-related signals~\cite{srirama2024hrp}, and gradually moved toward stronger spatial and geometric understanding through multi-view inputs, 3D structure, and explicit geometric modeling~\cite{hou20254d,jia2024lift3d,liu2024robouniview,seo2023multi}. More recent studies further incorporate actions, dynamics, or robot interaction data into pretraining, pushing representation learning beyond static visual encoding~\cite{afro,zhu2025lava,tian2026dynarend,yang2024spatiotemporal,zeng2024learning}. However, most existing methods are still used in downstream policies as single-frame encoders, and cannot provide long-horizon history in a compact form. RoboAct-CLIP~\cite{zhang2025roboact} starts to encode video, but still requires many tokens at inference and does not explicitly model action-state causality. In contrast, our work learns to compress vision-action history into a single compact final-token representation during pretraining, providing dense temporal representations to downstream policies with minimal interface cost.

\textbf{Memory Modeling for Robot Manipulation.}
Robot manipulation is inherently sequential and often partially observable, making memory a central issue in policy learning. The simplest solution is to stack historical observations or features, but this increases computation, memory usage, and inference latency, while leaving temporal abstraction to the policy itself. Existing memory methods mainly fall into three categories: compressing history before feeding it to the policy~\cite{wang2025lola,chen2025history,wei2026cyclemanip,jang2025contextvla}, maintaining an external memory bank for retrieval without changing the main policy structure~\cite{shi2025memoryvla,lin2025echovla,koo2025hamlet}, and exploiting the built-in temporal states of specific models, such as SAM2-style~\cite{fang2025sam2act} state mechanisms or Mamba latent states. Some recent works also study memory in VLA-style systems by attaching lightweight memory modules or retrieval mechanisms while preserving the original backbone~\cite{chen2026rmbench}. Overall, these methods focus on how memory is used at the \emph{policy stage}. Our work instead studies whether memory can be learned \emph{during pretraining}. Specifically, we pretrain a compact history representation from long-horizon vision-action interaction history and subsequently use it as a transferable temporal representation for downstream robot manipulation tasks.

\section{Preliminary}

\textbf{Mamba.}
Mamba is a selective state space model for sequence modeling. Given an input sequence $\mathbf{x}=[x_1,\dots,x_T]$, it maintains an internal state that evolves recurrently,
\begin{equation}
s_t = \mathcal{F}(s_{t-1}, x_t), \qquad
h_t = \mathcal{G}(s_t),
\end{equation}
where $\mathcal{F}$ denotes the Mamba state transition and $\mathcal{G}$ maps the internal state to the contextual output $h_t$. Each output $h_t$ therefore summarizes the current token and its preceding context, making Mamba suitable for long-horizon interaction history. In our method, we do not expose the internal recurrent state or append a dedicated memory token; instead, we use the Mamba-encoded output of the final vision token as a compact sequence-level history representation.

\textbf{Masked autoencoding.}
Masked autoencoding learns representations by reconstructing missing content from partially observed inputs. Let $\mathcal{M}$ be the set of masked positions. The encoder processes visible tokens, while the decoder predicts the masked content, with the objective
\begin{equation}
\mathcal{L}_{\mathrm{MAE}}
= \sum_{i \in \mathcal{M}} \ell(\hat{x}_i, x_i),
\end{equation}
where $x_i$ is the ground-truth token at masked position $i$, and $\hat{x}_i$ is the decoder reconstruction. Unlike direct compression, this objective encourages the latent representation to capture token dependencies. In our setting, this is important because useful memory should model not only what happened, but also how actions drive subsequent state changes.

\section{Method}

\begin{figure*}[t]
    \centering
    \includegraphics[width=\textwidth]{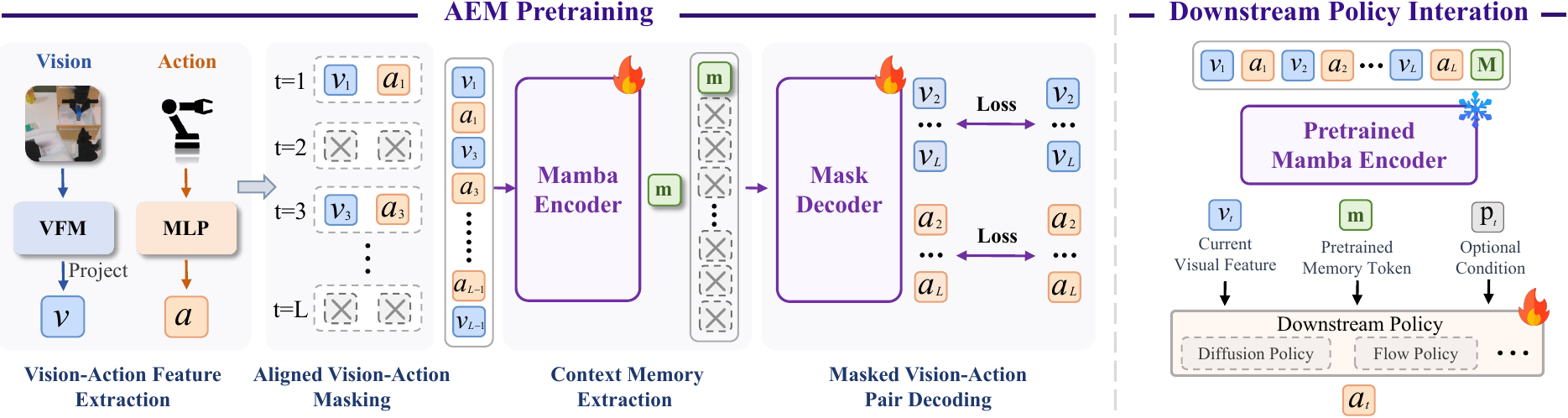}
    \caption{
    Overview of the proposed \textbf{AEM} framework. 
    During pretraining, visual observations and robot actions are first projected into a shared token space and organized as interleaved vision-action pairs. 
    AEM then applies aligned masking at the time-step level, where the visual and action tokens of the same step are masked together. 
    The visible tokens are fed into a Mamba encoder, and the encoded output of the final vision token serves as a compact single-vector history representation. 
    A lightweight Transformer decoder reconstructs the masked vision-action pairs using this final-token representation as global context, supervised by the reconstruction loss. 
    After pretraining, the learned encoder compresses the full vision-action history into the encoded final vision token, which is injected into downstream policies together with the current visual feature and optional task condition.
    }
    \label{fig:aem_framework}
\end{figure*}

\subsection{Overview}

We propose \textbf{AEM}, an Action-Effect Memory pretraining framework for robot manipulation in this work. The key idea is simple: instead of giving the policy raw history or learning memory only at the policy stage, we pretrain a compact temporal representation from long-horizon interaction history and inject it into downstream policies as temporal context. This turns historical context into a dense and reusable feature rather than a growing list of input tokens, without introducing a dedicated memory token.

We directly use visual features extracted by a visual foundation model, such as DINOv2~\cite{oquab2023dinov2}. Let $v_t \in \mathbb{R}^{D}$ denote the visual feature at time step $t$, and let $a_t \in \mathbb{R}^{d_a}$ denote the robot action. We use vision-action history, rather than visual history alone, because manipulation is fundamentally an action-driven process: the world changes not only because time passes, but because the robot acts on it. Therefore, a useful history representation should encode both what was observed and what action caused the subsequent change.

Given a history window of length $L$, \textbf{AEM} learns a compact history representation
\begin{equation}
m = g\!\left((v_1,a_1),\dots,(v_L,a_L)\right),
\end{equation}
which summarizes the interaction history and can be transferred to downstream policies. The overall framework is shown in Fig.~\ref{fig:aem_framework}, and the compact pretraining and downstream usage procedure is summarized in Algorithm~\ref{alg:aem}.

\begin{algorithm}[t]
\caption{\textcolor{AEMPurple}{AEM: Final-Vision-Token Pretraining}}
\label{alg:aem}
\small

\KwInput{
History $\{(\vis{v_t},\act{a_t})\}_{t=1}^{L}$
}

\BlankLine
\stage{A}{Pretraining}

\begin{enumerate}[leftmargin=1.5em, itemsep=0.15em, topsep=0.2em]
    \item Project and interleave:
    $\mathbf{x}=[\vis{z_1^v},\act{z_1^a},\ldots,\vis{z_L^v},\act{z_L^a}]$.

    \item Sample masked steps $\mathcal{M}$ while keeping the final visual token visible, and mask each selected pair
    $(\vis{z_t^v},\act{z_t^a})$.

    \item Encode visible tokens:
    $\mathbf{h}=\mathrm{Enc}_{\mathrm{Mamba}}(\mathbf{x}_{\backslash\mathcal{M}})$.

    \item Extract compact history state from the final vision token:
    $\mem{m}=\mathbf{h}_{L}^{v}$.

    \item Decode masked pairs from $\mem{m}$ and optimize
    $\mathcal{L}_{\mathrm{AEM}}=\lambda_v\mathcal{L}_{v}+\lambda_a\mathcal{L}_{a}$.
\end{enumerate}

\BlankLine
\stage{B}{Downstream Policy Integration}

\begin{enumerate}[leftmargin=1.5em, itemsep=0.15em, topsep=0.2em]
    \item Encode full history and extract the encoded final vision token:
    $\mem{m_t}=\mathrm{Enc}_{\mathrm{AEM}}(\{(\vis{v_i},\act{a_i})\}_{i<t})_{v,\mathrm{last}}$.

    \item Predict action:
    $\act{a_t}=\pi(\vis{v_t},\mem{m_t},p_t)$.
\end{enumerate}

\end{algorithm}

\subsection{AEM Pretraining}

\textbf{Vision-action sequence construction.}
We project visual and action features into a shared latent space using a visual projector $P_v$ and an action projector $P_a$, and then interleave them in temporal order to form a vision-action sequence. We further add learnable type embeddings and relative-time embeddings. This gives a sequence in which each time step is represented by a coupled visual-action token pair.

\textbf{Masked action-effect pretraining for compact history.}
During pretraining, we randomly sample a temporal window from the full trajectory, with variable window length and temporal stride. We then perform aligned masking at the time-step level: if the $t$-th step is masked, both its visual and action tokens are masked together. This preserves the interaction unit and encourages the model to reason about action-state dependencies rather than isolated tokens.

Only visible tokens are fed into the Mamba encoder. To impose a single-vector bottleneck without adding a dedicated memory token, we keep the final visual token visible and use it as the compression anchor:
\begin{equation}
\mathbf{h} = \mathrm{Enc}_{\mathrm{Mamba}}
\left(\mathbf{x}_{\backslash \mathcal{M}}\right),
\qquad
m = \mathbf{h}^{v}_{L},
\end{equation}
where $\mathbf{h}^{v}_{L}$ denotes the Mamba-encoded output at the final vision-token position. This compact final-token representation is the only encoder output passed to the decoder, forcing the model to compress the visible interaction history into a single latent representation.

To reconstruct masked tokens, we build a full-length decoder input using learnable mask tokens together with position, time, and type embeddings. The compact final-token representation $m$ is injected into every decoder position as the global context. A lightweight Transformer decoder then predicts both visual and action targets, where $Q_v(\cdot)$ and $Q_a(\cdot)$ are the output heads that map decoder features back to the visual-feature space and the action space, respectively.

We supervise only the masked positions during pretraining. For both branches, the reconstruction loss combines mean squared error and cosine distance:
\begin{equation}
\ell(\hat{y},y)
=
\|\hat{y}-y\|_2^2
+
\alpha \bigl(1-\cos(\hat{y},y)\bigr),
\end{equation}
where $\alpha$ is a fixed scalar. The overall objective is
\begin{equation}
\mathcal{L}_{\mathrm{AEM}}
=
\lambda_v
\sum_{t \in \mathcal{M}}
\ell(\hat{v}_t, v_t)
+
\lambda_a
\sum_{t \in \mathcal{M}}
\ell(\hat{a}_t, a_t).
\end{equation}
Here $\lambda_v$ and $\lambda_a$ are hyperparameters that balance the visual and action reconstruction terms.

\begin{figure*}[ht]
    \centering
    \includegraphics[width=0.9\textwidth]{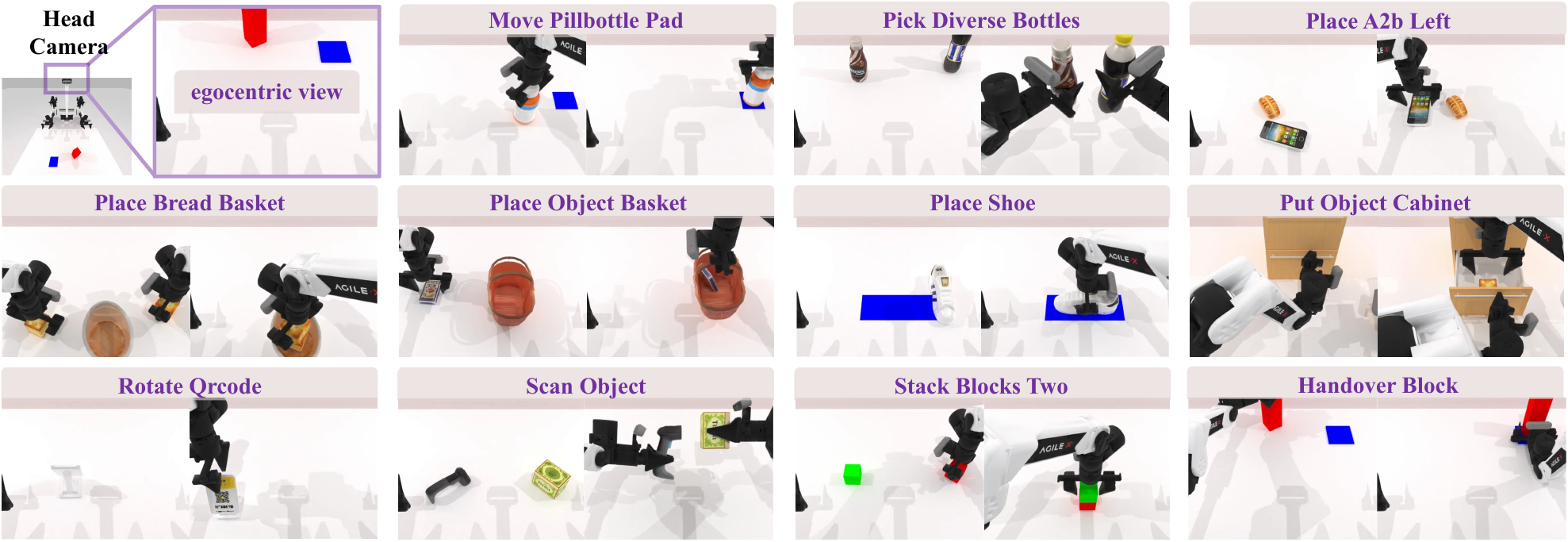}
    \caption{Simulation benchmark overview covering eleven diverse manipulation tasks,including picking, placing, pulling, rotating, handing over and so on.}
    \label{fig:sim_env_placeholder}
\end{figure*}

\textbf{Final-vision-token history extraction.}
After pretraining, \textbf{AEM} encodes the full interleaved history without masking. We use the Mamba-encoded output at the final vision-token position as the compact history representation,
\begin{equation}
m = \mathbf{h}^{v}_{L},
\end{equation}
which serves as a compact and dense summary of long-horizon interaction history for efficient control. Unlike token-preserving history encoders, \textbf{AEM} always produces a single vector regardless of the input history length, and it does so without adding a dedicated memory token.

\subsection{Downstream Policy Integration}

Let $p_t$ denote other policy conditions, such as language instructions, task identifiers, or goal specifications. A standard visuomotor policy predicts the action as
\begin{equation}
a_t = \pi(v_t, p_t).
\end{equation}
After introducing \textbf{AEM}, we use the pretrained final-token history representation together with the current feature:
\begin{equation}
a_t = \pi(v_t, m, p_t).
\end{equation}
In this way, the downstream policy receives temporal context at each step through a single pretrained history representation rather than an explicit stack of historical observations or a dedicated memory token.

\subsection{Pre-training Details}

For pretraining, we use DINOv2 CLS features ($D=768$) and 14-D robot actions as vision-action inputs. Both modalities are projected to a 512-D latent space and encoded by a 6-layer Mamba encoder with state dimension 256; a lightweight 2-layer decoder reconstructs masked visual and action targets. We sample 32-step windows without frame skipping and apply 70\% aligned time-step masking while keeping the final visual token visible as the compression anchor. AEM is trained with AdamW, learning rate $1.0\times10^{-5}$, weight decay 0.05, batch size 128, EMA 0.99, and 200 epochs. The reconstruction loss uses MSE plus cosine distance with weight 0.1.

\section{Simulation Experiment}

\subsection{Experimental Setup}

\textbf{Simulation benchmark.}
We evaluate AEM on standard manipulation and memory-dependent manipulation. For standard manipulation, we use eleven RoboTwin2.0~\cite{chen2025robotwin} tasks covering handover, placement, stacking, cabinet interaction, scanning, and object reorientation. For memory-dependent evaluation, we use RMBench~\cite{chen2026rmbench} Put Back Block and Swap Blocks, where the current observation is insufficient and the policy must retain task-relevant history. RoboTwin2.0 is evaluated under clean scenes and randomized scenes with distractors, pose changes, lighting variation, and texture changes.

\textbf{Baseline methods.}
We test whether AEM can improve different policy families without changing their action-generation backbone. We use Diffusion Policy (DP)~\cite{chi2025diffusion} and ManiFlow~\cite{yan2025maniflow} as baselines. DP predicts action chunks through a 100-step DDPM denoising process, while ManiFlow is a flow-matching policy. For AEM variants, the pretrained history representation is concatenated with the current visual feature and fed to the same policy backbone.

\textbf{Implementation details.}
All downstream policies use single-frame observations, while AEM injects history through one compact vector. For DP, the action horizon is 8, the executed chunk is 6, and the AEM history window is 16. DP uses a ResNet visual backbone, U-Net channels $[256,512,1024]$, DDPM with a squared-cosine beta schedule, and epsilon prediction. For ManiFlow, we use its R3M-initialized visual encoder and fine-tune it with the policy. All policies use $224\times224$ images, batch size 128, 50 warmup steps, and 1000 epochs. Training uses AdamW, learning rate $1.0\times10^{-4}$, and cosine scheduling. We evaluate every 100 epochs with 25 rollouts and report the best success rate. All experiments run on a single NVIDIA RTX 4090 GPU.

\subsection{Result Comparison}

\begin{table*}[t]
\centering
\caption{Main simulation results on eleven RoboTwin2.0 manipulation tasks, comparing Diffusion Policy (DP) and ManiFlow with their AEM-enhanced variants.
Success rates are in \%; AEM-enhanced rows are bold with purple highlighting, and red values indicate absolute improvements over the corresponding baseline average.}
\label{tab:main_results}
\scriptsize
\setlength{\tabcolsep}{11pt}
\renewcommand{\arraystretch}{1.12}
\begin{tabular}{llcccccc}
\toprule
\textbf{Method}
& \textbf{Type}
& \makecell{\textbf{Handover}\\\textbf{Block}}
& \makecell{\textbf{Put Object}\\\textbf{Cabinet}}
& \makecell{\textbf{Place Object}\\\textbf{Basket}}
& \makecell{\textbf{Place}\\\textbf{Shoe}}
& \makecell{\textbf{Stack Blocks}\\\textbf{Two}}
& \makecell{\textbf{Place Bread}\\\textbf{Basket}} \\
\midrule
DP
& w/o History
& 36.0 & 32.0 & 52.0 & 40.0 & 4.0 & 16.0 \\
\rowcolor{AEMRow}[\tabcolsep][\tabcolsep]
DP+AEM
& w/ History
& \textbf{96.0} & \textbf{52.0} & \textbf{76.0} & \textbf{76.0} & \textbf{12.0} & \textbf{40.0} \\
ManiFlow
& w/o History
& 4.0 & 12.0 & 40.0 & 28.0 & 0.0 & 16.0 \\
\rowcolor{AEMRow}[\tabcolsep][\tabcolsep]
ManiFlow+AEM
& w/ History
& \textbf{44.0} & \textbf{48.0} & \textbf{48.0} & \textbf{52.0} & \textbf{12.0} & \textbf{20.0} \\
\midrule[0.8pt]
\textbf{Method}
& \textbf{Type}
& \makecell{\textbf{Move Pillbottle}\\\textbf{Pad}}
& \makecell{\textbf{Pick Diverse}\\\textbf{Bottles}}
& \makecell{\textbf{Place A2B}\\\textbf{Left}}
& \makecell{\textbf{Rotate}\\\textbf{QR Code}}
& \makecell{\textbf{Scan}\\\textbf{Object}}
& \textbf{Avg.} \\
\midrule
DP
& w/o History
& 12.0 & 60.0 & 8.0 & 32.0 & 36.0 & 29.8 \\
\rowcolor{AEMRow}[\tabcolsep][\tabcolsep]
DP+AEM
& w/ History
& \textbf{20.0} & \textbf{68.0} & \textbf{24.0} & \textbf{44.0} & \textbf{48.0} & \textbf{50.5} \gain{+20.7} \\
ManiFlow
& w/o History
& 0.0 & 0.0 & 0.0 & 8.0 & 0.0 & 9.8 \\
\rowcolor{AEMRow}[\tabcolsep][\tabcolsep]
ManiFlow+AEM
& w/ History
& \textbf{8.0} & \textbf{24.0} & \textbf{12.0} & \textbf{48.0} & \textbf{4.0} & \textbf{29.1} \gain{+19.3} \\
\bottomrule
\end{tabular}
\end{table*}

\begin{figure}[t]
    \centering
    \includegraphics[width=1\linewidth]{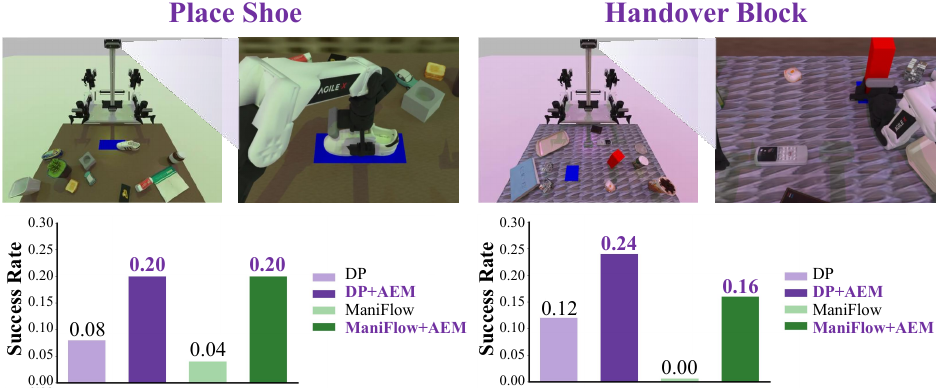}
    \caption{Randomized-scene evaluation. The figure includes environment examples and success-rate comparison for the tasks.}
    \label{fig:random_bar}
\end{figure}

\textbf{Does compact action-effect memory improve standard manipulation?}
To investigate whether a compact action-effect memory can improve standard manipulation, we evaluate AEM on eleven RoboTwin2.0 tasks, whose clean-task visualizations are shown in Fig.~\ref{fig:sim_env_placeholder}. These tasks cover diverse interaction patterns, including handover, placement, stacking, cabinet operation, scanning, and object reorientation, providing a broad testbed for standard dual-arm manipulation. Table~\ref{tab:main_results} compares DP and ManiFlow with and without AEM under the same action-generation backbones. AEM improves DP from \textbf{29.8\%} to \textbf{50.5\%} and ManiFlow from \textbf{9.8\%} to \textbf{29.1\%} on average. The consistent gains across two policy families suggest that the learned history representation is not tied to a specific policy architecture, but provides complementary temporal context that helps the policy infer task progress and interaction state beyond the current observation.

\begin{figure}[t]
    \centering
    \includegraphics[width=1\linewidth]{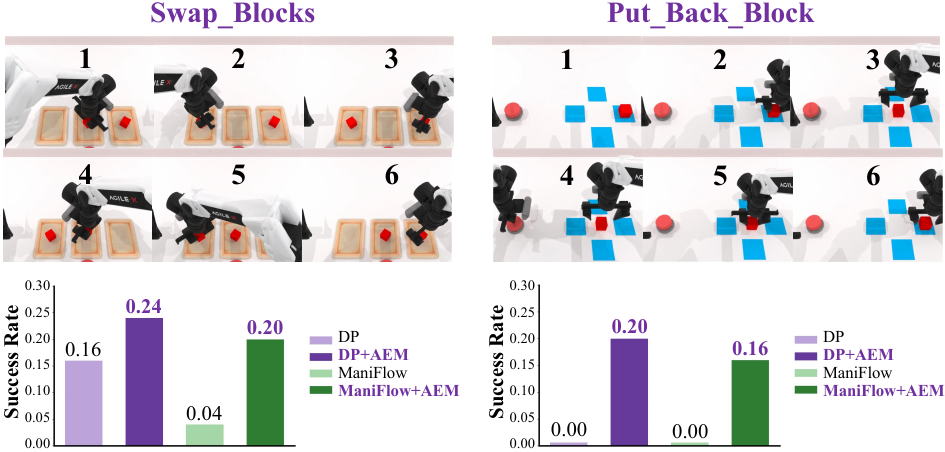}
    \caption{Memory-dependent RMBench evaluation on non-Markovian manipulation tasks. The figure compares DP and DP+AEM on tasks that require retaining task-relevant history.}
    \label{fig:rmbench}
\end{figure}

\textbf{Does AEM remain useful under randomized scenes?}
To further examine whether AEM remains effective when visual observations become less reliable, we select two representative tasks, Place Shoe and Handover Block, and evaluate them under randomized setting. The randomized setting introduces distractors, object-pose perturbations, lighting changes, and texture variation, which can make the current frame ambiguous or visually noisy. As shown in Fig.~\ref{fig:random_bar}, the baseline DP suffers a clear performance drop under randomization, whereas DP+AEM retains higher success rates on both selected tasks. This indicates that the compact history representation can provide additional temporal cues when the current observation alone is unstable, helping the policy better identify the manipulation phase and recover from visual perturbations.

\textbf{Is AEM more beneficial for explicitly non-Markovian tasks?}
To test whether AEM is especially useful when manipulation strongly depends on memory, we evaluate DP and DP+AEM on the RMBench Put Back Block and Swap Blocks tasks. Here, the current observation is insufficient; the policy must remember prior object configurations and interactions before deciding later placements. As shown in Fig.~\ref{fig:rmbench}, AEM brings larger improvements on these non-Markovian tasks than on standard manipulation tasks. This supports our hypothesis that masked action-effect pretraining condenses temporal information into a short final-token bottleneck. Thus, the compact memory is not just generic feature enhancement, but a task-relevant history representation especially valuable when control requires recalling past states and effects.

\begin{table*}[t]
\centering
\caption{
Efficiency and performance comparison with direct DINOv2 history stacking on Place Shoe.
Latency and FLOPs are measured per control step.
}
\label{tab:efficiency}
\scriptsize
\setlength{\tabcolsep}{10pt}
\renewcommand{\arraystretch}{1.12}
\begin{tabular}{lcccccccccc}
\toprule
\textbf{Method}
& \textbf{History}
& \makecell{\textbf{DINOv2}\\\textbf{/ G}}
& \makecell{\textbf{Mamba}\\\textbf{/ G}}
& \makecell{\textbf{Resnet}\\\textbf{/ G}}
& \makecell{\textbf{UNet}\\\textbf{/ G}}
& \makecell{\textbf{Total}\\\textbf{G}}
& \makecell{\textbf{Total}\\\textbf{ms}}
& \textbf{FPS}
& \makecell{\textbf{Clean}\\\textbf{Succ.}}
& \makecell{\textbf{Random}\\\textbf{Succ.}} \\
\midrule
DP 
& 1 frame 
& -- & -- & 2.37 & 15.80 & 18.20 & 623.30 & 1.60 & 0.40 & 0.08 \\
DINOv2 Concat 
& 1 frame 
& 22.00 & -- & 2.37 & 16.90 & 41.20 & 752.37 & 1.33 & 0.52 & 0.12 \\
DINOv2 Stack 
& 8 frames 
& 22.00 & -- & 2.37 & 24.60 & 48.90 & 796.94 & 1.25 & 0.52 & 0.12 \\
DINOv2 Stack 
& 16 frames 
& 22.00 & -- & 2.37 & 33.40 & 57.80 & 824.21 & 1.21 & 0.40 & 0.12 \\
\rowcolor{AEMRow}[\tabcolsep][\tabcolsep]
DP+AEM 
& 16-step memory 
& 22.00 & 0.80 & 2.37 & 16.50 & 41.70 & 730.01 & 1.37 & \best{0.76} & \best{0.20} \\
\bottomrule
\end{tabular}
\end{table*}

\textbf{Is compact memory better than direct history stacking?}
Table~\ref{tab:efficiency} compares AEM with direct DINOv2 feature concatenation and stacking on Place Shoe. We separate the cost of visual encoding, memory encoding, and diffusion denoising. A direct one-frame DINOv2 concatenation baseline tests whether stronger pretrained visual features alone explain the gain. The result shows that DINOv2 features are not sufficient: direct concatenation is weaker than AEM, while stacking more DINOv2 frames increases latency and FLOPs without improving success. This supports our claim that history should be compressed into a compact action-effect representation rather than appended as redundant feature vectors.

\begin{figure}[t]
    \centering
    \includegraphics[width=1\linewidth]{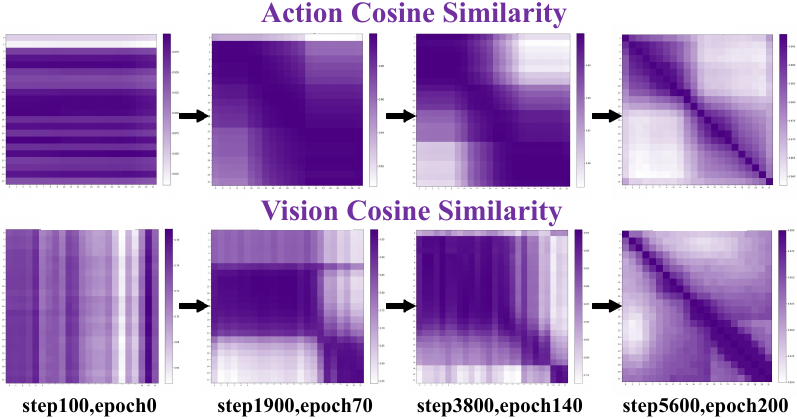}
    \caption{As training progresses, the single final token decodes increasingly accurate visual and action information.}
    \label{fig:similarity_matrix_placeholder}
\end{figure}

\textbf{Can a single final token preserve temporal information?}
We visualize whether the final-token bottleneck still preserves position-specific temporal information. On held-out trajectories, we mask a batch of visual features, decode them from the AEM memory, and compute cosine similarity between decoded and ground-truth features. A strong diagonal in Fig.~\ref{fig:similarity_matrix_placeholder} indicates that the compact final token preserves structured information about different temporal positions.

\subsection{Ablation Studies}

\textbf{How should AEM memory be fused, pretrained, and used?}
We ablate AEM design choices on Handover Block and Place Object Basket. Our full setting concatenates pretrained AEM memory with the current visual feature, pretrains on both observation and action, and feeds memory jointly with the current observation. Table~\ref{tab:ablation_design} compares it with four variants: \emph{Feature Add} sums rather than concatenates features; \emph{Joint Mamba} trains the memory encoder with the policy without AEM pretraining; \emph{Only Memory} removes the current observation; and \emph{Obs Only} pretrains without action reconstruction. Concatenation outperforms feature addition (0.96/0.76 vs.\ 0.88/0.52), while joint training (0.60), memory-only control (0.44/0.36), and observation-only pretraining (0.68/0.44) all fall short. AEM should therefore serve as a pretrained temporal complement to current perception, rather than replacing it or being learned solely during policy training.

\begin{table}[t]
\centering
\caption{
Fusion, pretraining, and memory-usage ablations.
Numbers denote success rates.
}
\label{tab:ablation_design}
\scriptsize
\renewcommand{\arraystretch}{1.12}
\setlength{\tabcolsep}{5.5pt}
\begin{tabular}{lccccc}
\toprule
\textbf{Task}
& \makecell{\textbf{Feature}\\\textbf{Add}}
& \makecell{\textbf{Joint}\\\textbf{Mamba}}
& \makecell{\textbf{Only}\\\textbf{Memory}}
& \makecell{\textbf{Obs}\\\textbf{Only}}
& \cellcolor{AEMRow}\textbf{Ours} \\
\midrule
Handover Block & 0.88 & 0.60 & 0.44 & 0.68 & \cellcolor{AEMRow}\best{0.96} \\
Place Object Basket & 0.52 & 0.60 & 0.36 & 0.44 & \cellcolor{AEMRow}\best{0.76} \\
\bottomrule
\end{tabular}
\end{table}

\begin{table}[t]
\centering
\caption{
Visual encoder ablation for AEM pretraining.
Numbers denote success rates.
}
\label{tab:ablation_encoder}
\scriptsize
\renewcommand{\arraystretch}{1.12}
\setlength{\tabcolsep}{4pt} 
\begin{tabular}{lccccc}
\toprule
\textbf{Task}
& \makecell{\textbf{DINOv2}\\\textbf{Pool}}
& \cellcolor{AEMRow}\makecell{\textbf{DINOv2}\\\textbf{CLS}}
& \makecell{\textbf{DINOv3}\\\textbf{Pool}}
& \makecell{\textbf{DINOv3}\\\textbf{CLS}}
& \makecell{\textbf{CLIP}\\\textbf{CLS}} \\
\midrule
Handover Block & 0.84 & \cellcolor{AEMRow}\best{0.96} & 0.72 & 0.76 & 0.84 \\
Place Object Basket & 0.48 & \cellcolor{AEMRow}\best{0.76} & 0.40 & 0.52 & 0.44 \\
\bottomrule
\end{tabular}
\end{table}

\textbf{Which visual encoder best supports compact memory learning?}
Table~\ref{tab:ablation_encoder} compares DINOv2~\cite{oquab2023dinov2}, DINOv3~\cite{simeoni2025dinov3}, and CLIP~\cite{radford2021learning} with pooled or CLS features. Encoder choice strongly affects compact memory quality: DINOv2 CLS performs best (0.96/0.76), and CLS tokens generally outperform pooling within each DINO variant, likely because pooled features mix object and background regions into a less stable compression target.

\begin{figure*}[t]
    \centering
    \includegraphics[width=\textwidth]{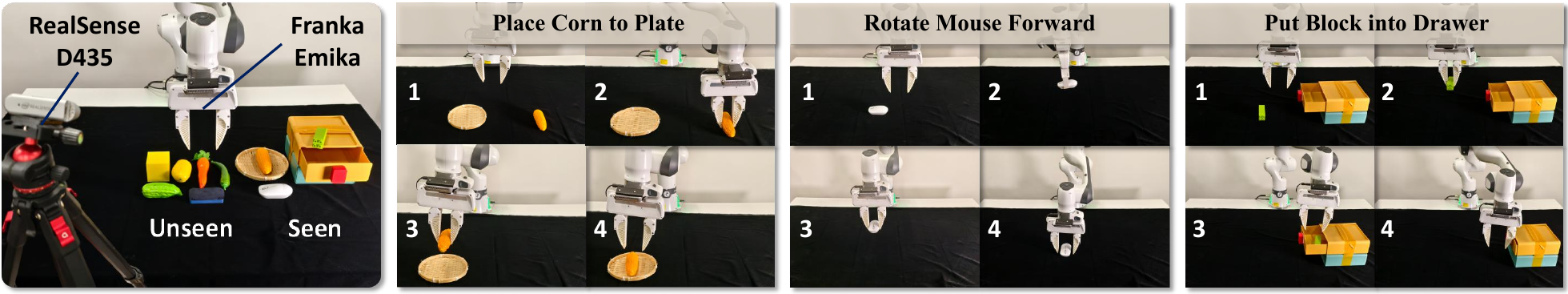}
    \caption{
    Real-world setup with a Franka Emika arm and an exocentric RealSense D435 view (left), together with demonstrations of the three evaluation tasks.
    }
    \label{fig:real-setup}
\end{figure*}

\begin{figure}[t]
    \centering
     \includegraphics[width=1\linewidth]{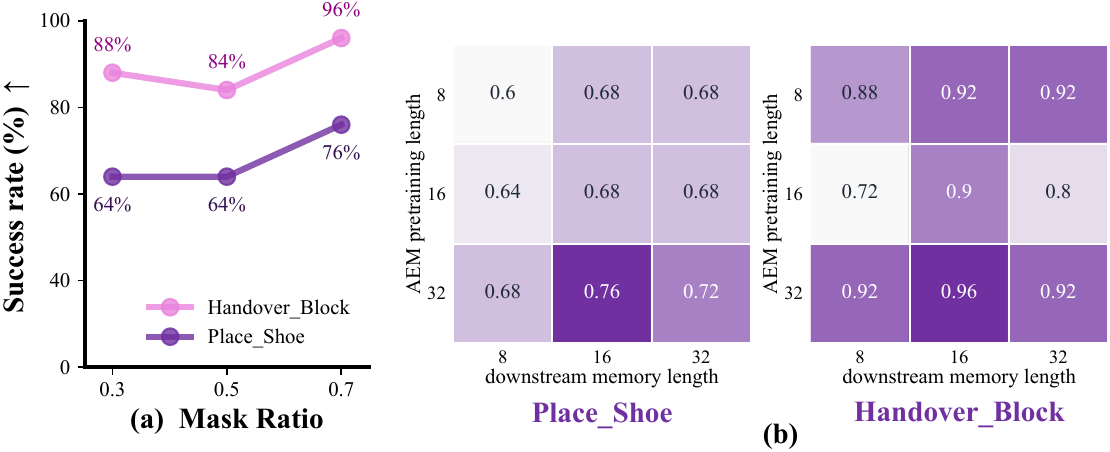}
    \caption{(a) Effect of mask ratio on downstream success for Handover Block and Place Object Basket. (b) Interaction between AEM pretraining length(y-axis) and downstream memory length(y-axis). The best setting uses a 32-step pretraining window and a 16-step inference window.}
    \label{fig:combine}
\end{figure}

\textbf{How much masking is needed for effective pretraining?}
We vary the aligned time-step mask ratio among 0.3, 0.5, and 0.7. As shown in Fig.~\ref{fig:combine} (a), downstream success improves with higher masking on both tasks, and 0.7 performs best, suggesting that harder reconstruction encourages the final token to encode longer-range temporal dependencies.

\textbf{How should pretraining and inference history lengths be matched?}
We study how the pretraining window and downstream memory horizon should be aligned. Fig.~\ref{fig:combine} (b) shows that a 32-frame pretraining window with a 16-step inference horizon works best, balancing temporal compression with a compact history signal for policy conditioning.

\section{Real-World Experiment}

\subsection{Experimental Setup}

\textbf{Does AEM transfer to physical manipulation?}
We evaluate AEM on a Franka Emika arm with an exocentric RealSense D435 camera, using the same DP and DP+AEM comparison as in simulation. AEM is pretrained on real-robot demonstrations from this platform and kept frozen during downstream policy training. As shown in Fig.~\ref{fig:real-setup}, we test three tabletop tasks: \emph{Place Corn to Plate}, where the arm picks up a corn cob and places it on a wooden plate; \emph{Rotate Mouse Forward}, where the arm grasps a sideways mouse, rotates it to face outward relative to the table edge, and puts it back down; and \emph{Put Block into Drawer }, where the arm places a block into a small drawer unit and then pushes the drawer door closed.

\subsection{Real-World Performance}

\textbf{Does AEM improve real-world task success?}
Table~\ref{tab:real_results} reports success rates under the standard tabletop setting. AEM improves all three tasks and raises the average success rate from 43.3\% to 81.7\%. The gains are most noticeable on \emph{Rotate Mouse Forward} and \emph{Put Block into Drawer }, where reorientation and drawer closure require tracking short-term interaction history beyond the current frame.

\textbf{Does AEM remain useful with task-irrelevant objects?}
We place additional objects on the table unrelated to the target task, including fruits, a sponge, and an extra block, while keeping the task targets unchanged. As reported in Table~\ref{tab:real_results}, both methods degrade under this cluttered setting, but DP+AEM retains higher success rates than DP on all three tasks (15\% vs.\ 68.3\% on average). The drop is especially clear on \emph{Place Corn to Plate}, where distractors near the plate make it easier for the baseline to lose the target after grasping. 

\begin{table}[t]
\centering
\caption{
Real-world success rates (\%) under the standard setting and with task-irrelevant distractor objects.
}
\label{tab:real_results}
\scriptsize
\setlength{\tabcolsep}{6pt}
\renewcommand{\arraystretch}{1.12}
\begin{tabular}{lcccc}
\toprule
\multirow{2}{*}{\textbf{Task}}
& \multicolumn{2}{c}{\textbf{Standard}}
& \multicolumn{2}{c}{\textbf{w/ Distractors}} \\
\cmidrule(lr){2-3}\cmidrule(lr){4-5}
& \textbf{DP} & \textbf{DP+AEM}
& \textbf{DP} & \textbf{DP+AEM} \\
\midrule
Place Corn to Plate & 60.0 & \best{95.0} & 25.0 & \best{85.0} \\
Rotate Mouse Forward & 50.0 & \best{80.0} & 20.0 & \best{75.0} \\
Put Block into Drawer & 20.0 & \best{70.0} & 0.0 & \best{45.0} \\
\midrule
Average & 43.3 & \best{81.7} & 15 & \best{68.3} \\
\bottomrule
\end{tabular}
\end{table}

\section{Conclusion}

Most robot representation pretraining treats manipulation as single-timestep visual encoding, overlooking temporal interaction structure. To address this, we proposed AEM, an Action-Effect Memory framework that compresses interleaved vision-action histories into a final-token representation via masked action-effect reconstruction. This fixed-size temporal bottleneck can be directly injected into existing visuomotor policies. Experiments on RoboTwin2.0, RMBench, and real robots show that AEM consistently improves Diffusion Policy and ManiFlow, especially under randomized scenes and non-Markovian tasks, while being more effective and efficient than direct DINOv2 history stacking.

A current limitation is the lack of large-scale pretraining validation. Future work will scale AEM to broader robot datasets and further strengthen the bottleneck to better distill task progress and action-consequence cues for control.

\section{Acknowledgments}

The computation and experimental validation in this paper were proudly sponsored and supported by Beijing Suanli Ziyou Technology Co., Ltd. (\href{https://www.gpufree.cn}{gpufree.cn}). Thanks to the stable GPU resources they provided, the complex neural network pre-training and embodied task evaluations involved in this study were completed efficiently. We extend our deepest appreciation for their support.

\bibliographystyle{IEEEtran}
\bibliography{reference}

@article{ma2022vip,
  title={Vip: Towards universal visual reward and representation via value-implicit pre-training},
  author={Ma, Yecheng Jason and Sodhani, Shagun and Jayaraman, Dinesh and Bastani, Osbert and Kumar, Vikash and Zhang, Amy},
  journal={arXiv preprint arXiv:2210.00030},
  year={2022}
}

@inproceedings{afro,
  title={Bootstrap Dynamic-Aware 3D Visual Representation for Scalable Robot Learning},
  author={Liang, Qiwei and Cai, Boyang and Lai, Minghao and Zhuang, Sitong and Lin, Tao and Qin, Yan and Ye, Yixuan and Liang, Jiaming and Xu, Renjing},
  booktitle={Proceedings of the IEEE/CVF Conference on Computer Vision and Pattern Recognition},
  pages={13419--13429},
  year={2026}
}

@inproceedings{r3m_ref,
  title={R3M: A Universal Visual Representation for Robot Manipulation},
  author={Nair, Suraj and Rajeswaran, Aravind and Kumar, Vikash and Finn, Chelsea and Gupta, Abhinav},
  booktitle={Conference on Robot Learning (CoRL)},
  pages={892--909},
  year={2023},
}

@inproceedings{3d-mvp,
  title={3D-MVP: 3D Multiview Pretraining for Manipulation},
  author={Qian, Shengyi and Mo, Kaichun and Blukis, Valts and Fouhey, David F and Fox, Dieter and Goyal, Ankit},
  booktitle={Proceedings of the Computer Vision and Pattern Recognition Conference (CVPR)},
  pages={22530--22539},
  year={2025}
}

@article{mvp_ref,
  title={Masked Visual Pre-training for Motor Control},
  author={Tete Xiao and Ilija Radosavovic and Trevor Darrell and Jitendra Malik},
  journal={arXiv preprint arXiv:2203.06173},
  year={2022}
}

@article{zhang2025roboact,
  title={RoboAct-CLIP: Video-Driven Pre-training of Atomic Action Understanding for Robotics},
  author={Zhang, Zhiyuan and He, Yuxin and Sun, Yong and Shi, Junyu and Liu, Lijiang and Nie, Qiang},
  journal={arXiv preprint arXiv:2504.02069},
  year={2025}
}

@inproceedings{sensorimotor,
  title={Robot learning with sensorimotor pre-training},
  author={Radosavovic, Ilija and Shi, Baifeng and Fu, Letian and Goldberg, Ken and Darrell, Trevor and Malik, Jitendra},
  booktitle={Conference on Robot Learning (CoRL)},
  pages={683--693},
  year={2023},
}

@article{RPR,
  title={Robots pre-train robots: Manipulation-centric robotic representation from large-scale robot datasets},
  author={Jiang, Guangqi and Sun, Yifei and Huang, Tao and Li, Huanyu and Liang, Yongyuan and Xu, Huazhe},
  journal={arXiv preprint arXiv:2410.22325},
  year={2024}
}

@article{dynamo,
  title={Dynamo: In-domain dynamics pretraining for visuo-motor control},
  author={Cui, Zichen J and Pan, Hengkai and Iyer, Aadhithya and Haldar, Siddhant and Pinto, Lerrel},
  journal={Advances in Neural Information Processing Systems},
  pages={33933--33961},
  year={2024}
}

@inproceedings{spa,
  title={Spa: 3d spatial-awareness enables effective embodied representation},
  author={Zhu, Haoyi and Yang, Honghui and Wang, Yating and Yang, Jiange and Wang, Limin and He, Tong},
  booktitle={International Conference on Learning Representations},
  pages={26361--26391},
  year={2025}
}

@article{intelligence2026pi,
  title={{$\pi_{0.7}$: A Steerable Generalist Robotic Foundation Model with Emergent Capabilities}},
  author={Intelligence, Physical and Ai, Bo and Amin, Ali and Aniceto, Raichelle and Balakrishna, Ashwin and Balke, Greg and Black, Kevin and Bokinsky, George and Cao, Shihao and Charbonnier, Thomas and others},
  journal={arXiv preprint arXiv:2604.15483},
  year={2026}
}

@article{torne2026mem,
  title={Mem: Multi-scale embodied memory for vision language action models},
  author={Torne, Marcel and Pertsch, Karl and Walke, Homer and Vedder, Kyle and Nair, Suraj and Ichter, Brian and Ren, Allen Z and Wang, Haohuan and Tang, Jiaming and Stachowicz, Kyle and others},
  journal={arXiv preprint arXiv:2603.03596},
  year={2026}
}

@inproceedings{lin2026hif,
  title={Hif-vla: Hindsight, insight and foresight through motion representation for vision-language-action models},
  author={Lin, Minghui and Ding, Pengxiang and Wang, Shu and Zhuang, Zifeng and Liu, Yang and Tong, Xinyang and Song, Wenxuan and Lyu, Shangke and Huang, Siteng and Wang, Donglin},
  booktitle={Proceedings of the IEEE/CVF Conference on Computer Vision and Pattern Recognition},
  pages={20732--20742},
  year={2026}
}

@article{srirama2024hrp,
  title={Hrp: Human affordances for robotic pre-training},
  author={Srirama, Mohan Kumar and Dasari, Sudeep and Bahl, Shikhar and Gupta, Abhinav},
  journal={arXiv preprint arXiv:2407.18911},
  year={2024}
}

@article{vaswani2017attention,
  title={Attention is all you need},
  author={Vaswani, Ashish and Shazeer, Noam and Parmar, Niki and Uszkoreit, Jakob and Jones, Llion and Gomez, Aidan N and Kaiser, {\L}ukasz and Polosukhin, Illia},
  journal={Advances in neural information processing systems},
  year={2017}
}

@article{gu2023mamba,
  title={Mamba: Linear-time sequence modeling with selective state spaces},
  author={Gu, Albert and Dao, Tri},
  journal={arXiv preprint arXiv:2312.00752},
  year={2023}
}

@inproceedings{he2022masked,
  title={Masked autoencoders are scalable vision learners},
  author={He, Kaiming and Chen, Xinlei and Xie, Saining and Li, Yanghao and Doll{\'a}r, Piotr and Girshick, Ross},
  booktitle={Proceedings of the IEEE/CVF conference on computer vision and pattern recognition},
  pages={16000--16009},
  year={2022}
}

@article{chen2025robotwin,
  title={Robotwin 2.0: A scalable data generator and benchmark with strong domain randomization for robust bimanual robotic manipulation},
  author={Chen, Tianxing and Chen, Zanxin and Chen, Baijun and Cai, Zijian and Liu, Yibin and Li, Zixuan and Liang, Qiwei and Lin, Xianliang and Ge, Yiheng and Gu, Zhenyu and others},
  journal={arXiv preprint arXiv:2506.18088},
  year={2025}
}

@article{chen2026rmbench,
  title={Rmbench: Memory-dependent robotic manipulation benchmark with insights into policy design},
  author={Chen, Tianxing and Wang, Yuran and Li, Mingleyang and Qin, Yan and Shi, Hao and Li, Zixuan and Hu, Yifan and Zhang, Yingsheng and Wang, Kaixuan and Chen, Yue and others},
  journal={arXiv preprint arXiv:2603.01229},
  year={2026}
}

@article{chi2025diffusion,
  title={Diffusion policy: Visuomotor policy learning via action diffusion},
  author={Chi, Cheng and Xu, Zhenjia and Feng, Siyuan and Cousineau, Eric and Du, Yilun and Burchfiel, Benjamin and Tedrake, Russ and Song, Shuran},
  journal={The International Journal of Robotics Research},
  pages={1684--1704},
  year={2025},
}

@article{yan2025maniflow,
  title={Maniflow: A general robot manipulation policy via consistency flow training},
  author={Yan, Ge and Zhu, Jiyue and Deng, Yuquan and Yang, Shiqi and Qiu, Ri-Zhao and Cheng, Xuxin and Memmel, Marius and Krishna, Ranjay and Goyal, Ankit and Wang, Xiaolong and others},
  journal={arXiv preprint arXiv:2509.01819},
  year={2025}
}

@article{zhou2025mtil,
  title={Mtil: Encoding full history with mamba for temporal imitation learning},
  author={Zhou, Yulin and Lin, Yuankai and Peng, Fanzhe and Chen, Jiahui and Huang, Kaiji and Yang, Hua and Yin, Zhouping},
  journal={IEEE Robotics and Automation Letters},
  year={2025},
}

@inproceedings{hou20254d,
  title={4D Visual Pre-training for Robot Learning},
  author={Hou, Chengkai and Ze, Yanjie and Fu, Yankai and Gao, Zeyu and Hu, Songbo and Yu, Yue and Zhang, Shanghang and Xu, Huazhe},
  booktitle={Proceedings of the IEEE/CVF International Conference on Computer Vision},
  year={2025}
}

@inproceedings{radosavovic2023real,
  title={Real-world robot learning with masked visual pre-training},
  author={Radosavovic, Ilija and Xiao, Tete and James, Stephen and Abbeel, Pieter and Malik, Jitendra and Darrell, Trevor},
  booktitle={Conference on Robot Learning},
  year={2023},
}

@inproceedings{seo2023multi,
  title={Multi-view masked world models for visual robotic manipulation},
  author={Seo, Younggyo and Kim, Junsu and James, Stephen and Lee, Kimin and Shin, Jinwoo and Abbeel, Pieter},
  booktitle={International Conference on Machine Learning},
  year={2023},
}

@article{liu2024robouniview,
  title={Robouniview: Visual-language model with unified view representation for robotic manipulation},
  author={Liu, Fanfan and Yan, Feng and Zheng, Liming and Feng, Chengjian and Huang, Yiyang and Ma, Lin},
  journal={arXiv preprint arXiv:2406.18977},
  year={2024}
}

@article{jia2024lift3d,
  title={Lift3d foundation policy: Lifting 2d large-scale pretrained models for robust 3d robotic manipulation},
  author={Jia, Yueru and Liu, Jiaming and Chen, Sixiang and Gu, Chenyang and Wang, Zhilue and Luo, Longzan and Lee, Lily and Wang, Pengwei and Wang, Zhongyuan and Zhang, Renrui and others},
  journal={arXiv preprint arXiv:2411.18623},
  year={2024}
}

@article{zhu2025lava,
  title={LaVA-Man: Learning Visual Action Representations for Robot Manipulation},
  author={Zhu, Chaoran and Wang, Hengyi and Pang, Yik Lung and Oh, Changjae},
  journal={arXiv preprint arXiv:2508.19391},
  year={2025}
}

@article{tian2026dynarend,
  title={DynaRend: Learning 3D Dynamics via Masked Future Rendering for Robotic Manipulation},
  author={Tian, Jingyi and Wang, Le and Zhou, Sanping and Wang, Sen and Hua, Gang},
  journal={Advances in Neural Information Processing Systems},
  year={2026}
}

@article{zeng2024learning,
  title={Learning manipulation by predicting interaction},
  author={Zeng, Jia and Bu, Qingwen and Wang, Bangjun and Xia, Wenke and Chen, Li and Dong, Hao and Song, Haoming and Wang, Dong and Hu, Di and Luo, Ping and others},
  journal={arXiv preprint arXiv:2406.00439},
  year={2024}
}

@article{yang2024spatiotemporal,
  title={Spatiotemporal predictive pre-training for robotic motor control},
  author={Yang, Jiange and Liu, Bei and Fu, Jianlong and Pan, Bocheng and Wu, Gangshan and Wang, Limin},
  journal={arXiv preprint arXiv:2403.05304},
  year={2024}
}

@article{shang2024theia,
  title={Theia: Distilling diverse vision foundation models for robot learning},
  author={Shang, Jinghuan and Schmeckpeper, Karl and May, Brandon B and Minniti, Maria Vittoria and Kelestemur, Tarik and Watkins, David and Herlant, Laura},
  journal={arXiv preprint arXiv:2407.20179},
  year={2024}
}

@article{lepert2025masquerade,
  title={Masquerade: Learning from in-the-wild human videos using data-editing},
  author={Lepert, Marion and Fang, Jiaying and Bohg, Jeannette},
  journal={arXiv preprint arXiv:2508.09976},
  year={2025}
}

@article{fang2025sam2act,
  title={Sam2act: Integrating visual foundation model with a memory architecture for robotic manipulation},
  author={Fang, Haoquan and Grotz, Markus and Pumacay, Wilbert and Wang, Yi Ru and Fox, Dieter and Krishna, Ranjay and Duan, Jiafei},
  journal={arXiv preprint arXiv:2501.18564},
  year={2025}
}

@inproceedings{wei2026cyclemanip,
  title={CycleManip: Enabling Cycle-based Manipulation via Effective History Perception and Understanding},
  author={Wei, Yi-Lin and Liao, Haoran and Lin, Yuhao and Wang, Pengyue and Liang, Zhizhao and Liu, Guiliang and Zheng, Wei-Shi},
  booktitle={Proceedings of the IEEE/CVF Conference on Computer Vision and Pattern Recognition},
  pages={20780--20789},
  year={2026}
}

@article{chen2025history,
  title={History-Aware Visuomotor Policy Learning via Point Tracking},
  author={Chen, Jingjing and Fang, Hongjie and Wang, Chenxi and Wang, Shiquan and Lu, Cewu},
  journal={arXiv preprint arXiv:2509.17141},
  year={2025}
}

@article{jang2025contextvla,
  title={ContextVLA: Vision-Language-Action Model with Amortized Multi-Frame Context},
  author={Jang, Huiwon and Yu, Sihyun and Kwon, Heeseung and Jeon, Hojin and Seo, Younggyo and Shin, Jinwoo},
  journal={arXiv preprint arXiv:2510.04246},
  year={2025}
}

@article{wang2025lola,
  title={LoLA: Long Horizon Latent Action Learning for General Robot Manipulation},
  author={Wang, Xiaofan and Gao, Xingyu and Fu, Jianlong and Li, Zuolei and Fortier, Dean and Mullins, Galen and Kolobov, Andrey and Guo, Baining},
  journal={arXiv preprint arXiv:2512.20166},
  year={2025}
}

@article{shi2025memoryvla,
  title={Memoryvla: Perceptual-cognitive memory in vision-language-action models for robotic manipulation},
  author={Shi, Hao and Xie, Bin and Liu, Yingfei and Sun, Lin and Liu, Fengrong and Wang, Tiancai and Zhou, Erjin and Fan, Haoqiang and Zhang, Xiangyu and Huang, Gao},
  journal={arXiv preprint arXiv:2508.19236},
  year={2025}
}

@article{lin2025echovla,
  title={EchoVLA: Robotic Vision-Language-Action Model with Synergistic Declarative Memory for Mobile Manipulation},
  author={Lin, Min and Liang, Xiwen and Lin, Bingqian and Jingzhi, Liu and Jiao, Zijian and Li, Kehan and Ma, Yuhan and Liu, Yuecheng and Zhao, Shen and Zhuang, Yuzheng and others},
  journal={arXiv preprint arXiv:2511.18112},
  year={2025}
}

@article{koo2025hamlet,
  title={Hamlet: Switch your vision-language-action model into a history-aware policy},
  author={Koo, Myungkyu and Choi, Daewon and Kim, Taeyoung and Lee, Kyungmin and Kim, Changyeon and Seo, Younggyo and Shin, Jinwoo},
  journal={arXiv preprint arXiv:2510.00695},
  year={2025}
}

@article{oquab2023dinov2,
  title={Dinov2: Learning robust visual features without supervision},
  author={Oquab, Maxime and Darcet, Timoth{\'e}e and Moutakanni, Th{\'e}o and Vo, Huy and Szafraniec, Marc and Khalidov, Vasil and Fernandez, Pierre and Haziza, Daniel and Massa, Francisco and El-Nouby, Alaaeldin and others},
  journal={arXiv preprint arXiv:2304.07193},
  year={2023}
}

@article{simeoni2025dinov3,
  title={Dinov3},
  author={Sim{\'e}oni, Oriane and Vo, Huy V and Seitzer, Maximilian and Baldassarre, Federico and Oquab, Maxime and Jose, Cijo and Khalidov, Vasil and Szafraniec, Marc and Yi, Seungeun and Ramamonjisoa, Micha{\"e}l and others},
  journal={arXiv preprint arXiv:2508.10104},
  year={2025}
}

@inproceedings{radford2021learning,
  title={Learning transferable visual models from natural language supervision},
  author={Radford, Alec and Kim, Jong Wook and Hallacy, Chris and Ramesh, Aditya and Goh, Gabriel and Agarwal, Sandhini and Sastry, Girish and Askell, Amanda and Mishkin, Pamela and Clark, Jack and others},
  booktitle={International conference on machine learning},
  pages={8748--8763},
  year={2021},
}

\vfill

\end{document}